\def\year{2020}\relax
\documentclass[letterpaper]{article} 
\usepackage{aaai20}  
\usepackage{times}  
\usepackage{helvet} 
\usepackage{courier}  
\usepackage[hyphens]{url}  
\usepackage{graphicx} 
\def\UrlFont{\rm}  
\usepackage{graphicx}  
\usepackage{amssymb}
\usepackage{pifont}
\newcommand{\cmark}{\ding{51}}%
\newcommand{\xmark}{\ding{55}}%

\usepackage[dvipsnames]{xcolor}

\usepackage{amsmath}
\usepackage{xspace}
\usepackage{goodstuff}
\usepackage{ellipsis}
\usepackage[inline]{enumitem}

\usepackage{xcolor}

\newabbrev{\shop}{\textsc{Shop3}}
\newabbrev{\pred}{\textsc{Shop2}}

\title{Provenance-Based Assessment of Plans in Context}

\author{Scott E.~Friedman, Robert P.~Goldman, Richard G.~Freedman, Ugur Kuter, \\
\Large \textbf{Christopher Geib, Jeffrey Rye} \\
\{ sfriedman, rpgoldman, rfreedman, ukuter, cgeib, jrye \} @
  sift.net \\ SIFT, LLC \\ Minneapolis, MN, USA }

\begin{document}

\maketitle

\begin{abstract}
Many real-world planning domains involve diverse
information sources, external entities, and variable-reliability agents, all of which may impact the confidence, risk, and sensitivity of plans.
Humans reviewing a plan may lack context about these factors; however, this information is available during the domain generation, which means it can also be interwoven
into the planner and its resulting plans. 
This paper presents a provenance-based approach to explaining automated plans.
Our approach (1) extends the \shop HTN planner to generate dependency information, (2) transforms the dependency information into an established PROV-O representation, and (3) uses graph propagation and TMS-inspired algorithms to support dynamic and counter-factual assessment of information flow, confidence, and support.
We qualified our approach's explanatory scope with respect to explanation targets from the automated planning literature and the information analysis literature, and we demonstrate its ability to assess a plan's pertinence, sensitivity, risk, assumption support, diversity, and relative confidence.

\end{abstract}

\section{Introduction}
\label{sec:intro}

In complex, dynamic, and uncertain environments, it is critical that human operators understand machine-generated plans, including their sensitivity to world changes, their reliance on individual actors, their diversity of information sources, their core assumptions, and how risky they are. 
This paper contributes an approach to dynamically explain and explore machine-generated single- or multi-agent, single- or multi-goal plans using \emph{provenance-based} analysis and visualization strategies.

Most prior work on explainable planning focuses on inspecting algorithms (\ie{}, explicating the decision-making process), synchronizing mental models (\eg{}, because the user views the problem differently than the planner), and improving usability (\eg{}, making complex plans more interpretable) \cite{ChakrabortiEtAlIJCAI2020} and assumed fixed background domain knowledge.
In contrast, our provenance-based approach treats the plan as a tripartite dependency graph that helps explain the foundations, reliability, and sensitivity of the information that comprises the plan's states and actions.

\begin{figure}
  \begin{center}
  \includegraphics[width=\linewidth]{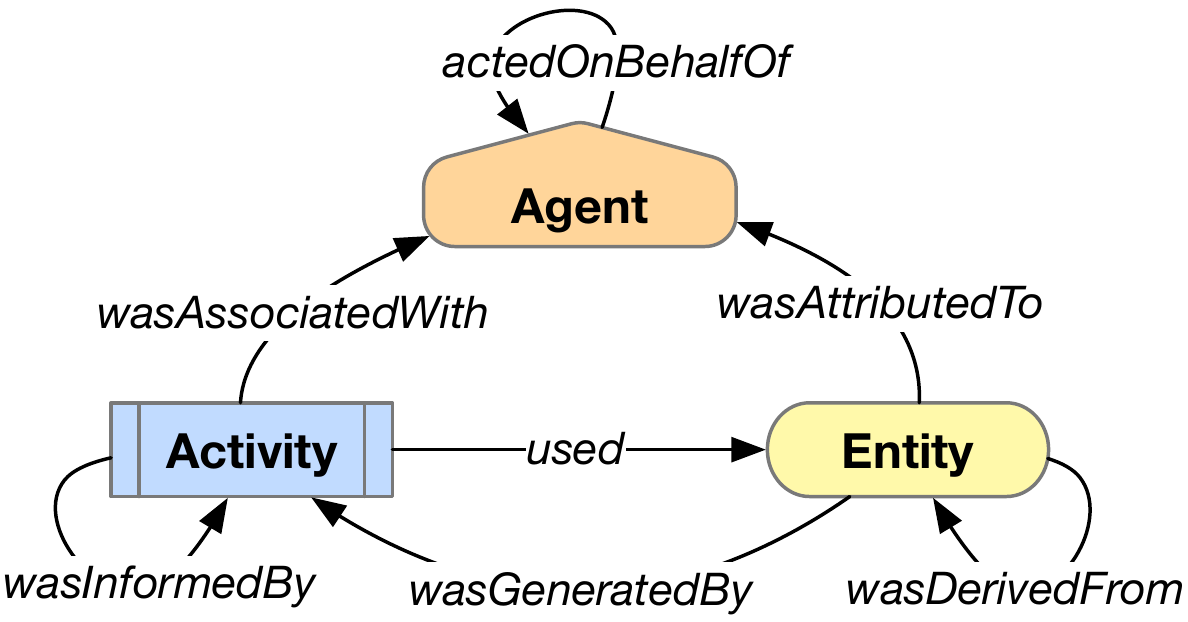}
  \caption{PROV ontology subset used in our approach.}
  \label{fig:prov}
  \end{center}
\end{figure}

We use the definition of ``provenance'' from the Provenance Data Model (PROV-DM): ``information about entities, activities, and people involved in producing a piece of data or thing, which can be used to form assessments about its quality, reliability or trustworthiness'' ~\cite{Moreau:13:PTP}.
We describe the formal PROV-DM relationships among these elements later, as shown in \figref{prov}.
We have augmented the Hierarchical Task Network (HTN) planner \shop{}~\cite{GoldmanKuter:SHOP3ELS} with the ability to annotate its plans with provenance by recording, on the fly,
\begin{enumerate*}[label=(\arabic*)]
\item causal dependencies,
\item dependencies from plan components onto aspects of the model (domain) from which they derive, and
\item sources of information used by the planner in checking preconditions and deriving beliefs.
\end{enumerate*}

The provenance of the \shop{} plan feeds into our downstream provenance analysis, which uses PROV-DM to represent beliefs, planned activities, and actors, and the recent DIVE ontology \cite{friedman_tapp_2020} to represent assumptions, confidence, and likelihood of those PROV-DM elements.
Our approach combines truth maintenance \cite{forbus1993building} and provenance propagation \cite{singh2018decision,gehani2010mendel,pasquier2016information} to estimate the confidence in the correctness of planned actions, and counterfactually assess the \textit{sensitivity} of the plan to [the absence of] various data sources, actors, events, and beliefs.

Our central claim is that tracking and analyzing a plan's provenance can improve the interpretation of plans--- along dimensions of confidence, information dependency, risk, and sensitivity--- without reducing the efficiency of the planner or the complexity of the search space.
To support this claim, we demonstrate our approach within a provenance visualization environment \cite{friedman_tapp_2020}.
This provenance-based approach is especially useful for explaining plans with multiple goals and for plans with multiple actions to achieve a given goal.
While our demonstration uses provenance analysis after planning completes, we identify future avenues for using provenance \emph{within} a planner to advise search heuristics, mixed-initiative planning, contingency planning, and replanning.


We continue with a review of relevant background in provenance-tracking and HTN planning.
We then describe our approach using provenance as a platform for plan explanation and assessment, qualifying the types of planning questions that our approach addresses. 
We demonstrate our system facilitating plan assessment, and we review the results and outline future work in our conclusion.




\section{Background}
\label{sec:background}

\subsection{Provenance-Tracking}
\label{sec:prov}

We utilize the PROV-O ontology \cite{lebo2013prov}, which expresses PROV Data Model's entities and relationships using the OWL2 Web Ontology Language.
The PROV Data Model includes the following three primary classes of elements to express provenance:
\begin{enumerate}
  \item \textbf{Entities} are real or hypothetical things with some fixed aspects in physical or conceptual space.  These may be beliefs, documents, databases, inferences, \etc
  \item \textbf{Activities} occur over a period of time, processing and/or generating entities.  These may be inference actions, judgment actions, planned (not yet performed) actions, \etc
  \item \textbf{Agents} are responsible for performing activities or generating entities.  These may be humans, machines, rovers, web services, \etc
\end{enumerate}

\noindent
The primary relationships over these three classes in PROV are shown in \figref{prov}, as detailed in the W3C PROV-O recommendation.\footnote{https://www.w3.org/TR/2013/REC-prov-o-20130430/}

Systems that utilize PROV-O, as specified in \figref{prov}, can represent long inferential chains, formally linking conclusions (\eg, a downstream belief) through generative activities (\eg, inference operations) and antecedents, to source entities and assumptions.
This comprises a directed network of provenance that we can traverse in either direction to answer questions of foundations, derivations, and impact.

\begin{figure}[t]
\centering
\includegraphics[width=\columnwidth]{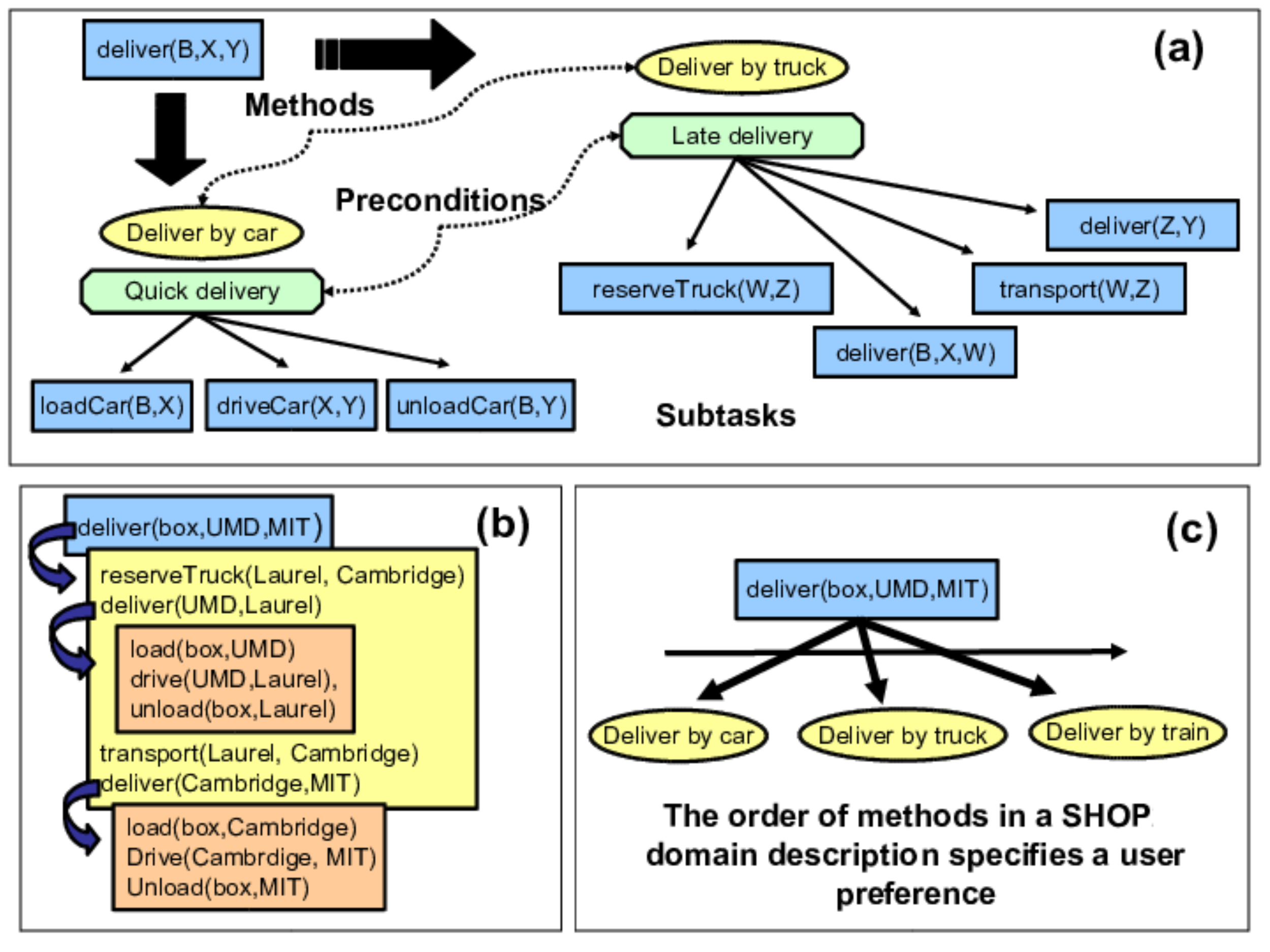} \vspace{-0.2in}
\caption{Delivery planning
example.} \label{fig:travel}
\end{figure}

\subsection{The DIVE Ontology}
\label{sec:dive}

The DIVE ontology \cite{friedman_tapp_2020} extends the PROV ontology with additional classes and relationships to appraise information and validate information workflows.
For this work, we use DIVE's \textbf{Appraisal} class, which is an \textbf{Agent}'s judgment about an activity, entity, or other agent.

For example, we express a DIVE \textbf{Appraisal} about a GPS sensor---from which we derive beliefs about the world before planning and during plan execution---with moderate baseline confidence.
This baseline confidence in our GPS sensor may affect our confidence of the information it emits, all else being equal, which may ultimately impact our judgment of the success likelihood of our planned actions.

We also use DIVE to express \textit{collection disciplines} such as GEOINT (geospatial), IMINT (image), and other types of information for all relevant information sources, beliefs, and sensors involved in a plan.
DIVE is expressed at the meta-level of PROV.
DIVE expressions flow through the network to facilitate downstream quality judgments and interpretation, as we demonstrate in this work.

\hide{
\subsection{Truth-Maintenance Systems}
\label{sec:tms}

Truth-Maintenance Systems (TMSs) \cite{forbus1993building,friedman2018csj} explicitly store entities alongside \emph{justifications} that link antecedent entities (analogous to PROV entities) with consequent entities.
This explicitly encodes the rationale for each entity, so --- similar to the PROV ontology --- we can use a TMS to explore foundations, derivations, and impact.

TMSs track \emph{environments} as sets of elements that sufficiently justify an entity in its upstream lineage.
If the lineage changes (e.g., due to a new derivation of an entity), the TMS recomputes the affected environments.
Environments allow TMSs to efficiently recognize contradictions, retrieve logical rationale, and identify upstream assumptions \cite{de1986assumption}.
TMSs operate alongside inference engines to record the lineage and logical conditions for believing various assertions; they do not themselves generate inferences or derive entities.
Our approach utilizes TMS-like environments to efficiently refute information, propagate confidence, and visualize impact.
}

\begin{figure*}
  \begin{center}
  \includegraphics[width=\textwidth]{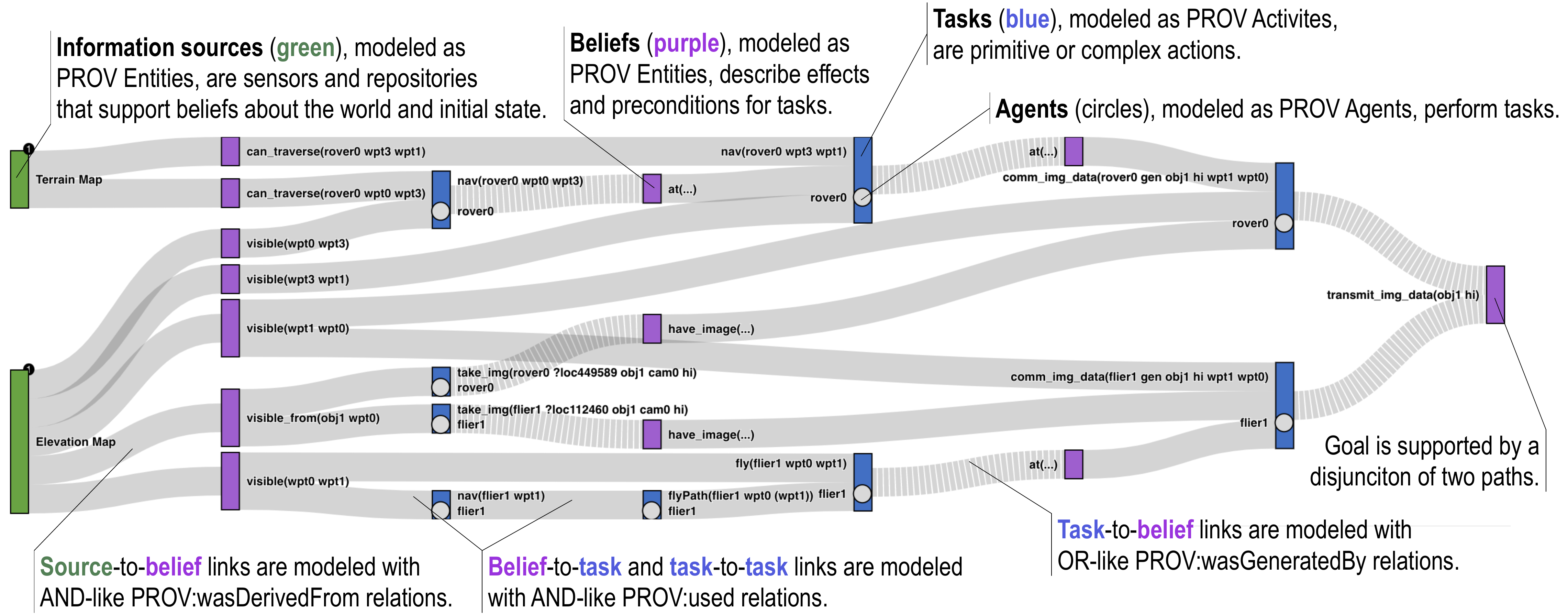}
  \caption{How we represent (with PROV) and display (with D3.js) the provenance of \shop plans.}
  \label{fig:prov-to-plan}
  \end{center}
\end{figure*}

\subsection{The SHOP3 HTN Planner}

\label{sec:shop3}

\shop~\cite{GoldmanKuter:SHOP3ELS} is the successor to the \pred HTN
planner~\cite{nau2003shop2} developed at the University of Maryland.
Unlike a first principles planner, an HTN planner produces a sequence of actions that perform some activity or
\emph{task}, instead of finding a path to a goal state.
An HTN planning domain includes a set of planning {\em operators}
(actions) and
\emph{methods}, each of which is a prescription for how to decompose a task into its
\emph{subtasks} (smaller tasks). The description of a planning problem contains an initial state as
in classical planning. Instead of a goal formula, however, there is a partially-ordered set of
tasks to accomplish.
Planning proceeds by decomposing tasks recursively into subtasks, until
\emph{primitive tasks}, which can be performed directly using the planning operators, are reached.
For each task, the planner chooses an applicable method, instantiates it to decompose the task into
subtasks, and then chooses and instantiates other methods to decompose the subtasks even further.
If the constraints on the subtasks or the interactions among them prevent the plan from being
feasible, the planner will backtrack and try other methods.
\figref{travel} illustrates how \shop HTN domains are described and used for planning in a Delivery planning example.

\shop is an HTN planner that generates actions in the
order they will be executed in the world (hence ``hierarchical
\emph{ordered} planner'' in the name).
\pred is relatively efficient and well-tested, and
it performed well in the 2002 IPC--the last IPC in which HTN planners competed~\cite{long03:_inter_plann_compet}.
\pred also has been used in a number of planning applications, including recently at SIFT for
Air Operations and UAV planning \cite{mueller17human},
cyber security, cyber-physical systems,
planning for synthetic biology experiments \cite{kuter18xplan}, and more.
For an earlier survey of SHOP2 applications,
see Nau, \etal~\shortcite{nau05applications}.
\shop retains the essential
features of \pred, but has a modernized codebase, is easier to extend
(\eg{} with plan repair capabilities, new input languages, \etc), and
an alternative, more efficient search engine.


\section{Approach}
\label{sec:approach}

We first describe how we extended the \shop planner to emit dependency information to support provenance.
We then describe our approach with respect to relevant questions from the planning literature \cite{MariaFoxExplainablePlanning2017} and information analysis literature \cite{icd_203,icd_206,zelik2010measuring} that have been proposed as primary targets for integrity and explainability.
We describe relevant representations and algorithms in our approach as they apply to these questions.

\subsection{\shop and Provenance Tracking}
\label{sec:planner-extensions}

In related work on plan repair~\cite{Goldman:StablePlanRepair:2020}, we have augmented \shop so that, when planning, it builds a plan tree that has dependency information (causal links).  These links allow the plan repair system to identify the minimally compromised subtree of the plan, as a way to provide \emph{stable}, minimal-perturbation plan repairs.  This extension provides much of the provenance information that we need for explainability, because it allows us to trace the choice of methods and primitive actions back to other choices that enabled them.
The present approach extends the scope and semantics of these links to
\begin{enumerate*}[label=(\arabic*)]
\item trace decisions back to the model components that justify them and
\item trace preconditions back to actions that establish them and information sources that provided them.
\end{enumerate*}

\textbf{Tracing decisions back to model components} is straightforward: the \shop planner takes as input \texttt{domain} and \texttt{problem} data structures, and the \texttt{domain} data structures contain the model components, specifically the primitive operator and method definitions.  For the moment, we do not track the provenance of components of the planner's model.
However, since the domain descriptions are typically maintained in a revision control system, such as subversion or git, it would be relatively simple to extend our provenance tracing back to the person or persons who wrote these model components.  For a more sophisticated development environment, one could imagine a traceback that reaches into an integrated development environment or a machine learning system.

\textbf{Tracing decisions back to information sources} is somewhat more difficult. In the base case, a proposition is established in the \texttt{problem} data structure -- that is, in the initial state.  In a larger system that incorporates \shop, there is generally a component that builds these \texttt{problem} data structures.  For example, in a robot planning system, we generally have a component that builds problems programmatically from user input (tasks to achieve) and some source of external information (\eg{}, a map database, telemetry from robotic platforms, \etc).  These components can annotate the initial state (and potentially the tasks \shop is asked to plan) with provenance information, using PROV-DM in a way that is appropriate to the application domain.  This provenance information can then be propagated through the causal links in the plan tree.

There is one remaining complication: in the interests of modeling efficiency and expressivity, \shop incorporates a theorem-prover -- a backward-chaining engine inspired by Prolog.
This is necessary because \shop{}'s expressive power is not limited to propositional logic, the way most planners are: it permits state axioms, and non-finite domains of quantification.\footnote{Note that though critical to \shop{} applications, these features must be used with care, because they can compromise soundness and completeness.} Thus some preconditions may be established not just causally, but inferentially, through Horn clause (``axiom'') deduction.  Accordingly, we must extend our theorem-prover so that it also provides traceability. Provenance annotations that traced provenance through axioms back to actions that established antecedents for the axioms were already in place for plan repair. These will now automatically incorporate information source provenance, as well as causal provenance.  At the moment, we do not trace the axioms themselves, but this would be a trivial extension.

\subsection{Mapping \shop plans to PROV}

Our system converts the extended \shop plans into the PROV data model, using the PROV-O ontology to represent the elements and relationships between them.
\figref{prov-to-plan} illustrates the SHOP-to-PROV mapping in a screenshot of our system displaying \shop planner output (some elements removed for simplicity).
The plan content in \figref{prov-to-plan} displays a single goal (at right) to transmit image data of \textbf{objective1} in high-resolution, and this goal is supported by two paths of tasks, performed by two separate agents (the aerial unit \textbf{flier1} and the land unit \textbf{rover0}), with foundational beliefs derived from a \textbf{Terrain Map} and an \textbf{Elevation Map}.
We use the following mapping:
\begin{itemize}
  \item \textbf{Planned Tasks} are specializations of \textbf{prov:Activity}.  Unlike traditional uses of provenance for tracking \textit{past} events, the PROV Activities from the plan may not yet have--or may never actually--occur.
  \item \textbf{Plan Actors} are specializations of \textbf{prov:Agent}.
  They are the performers of the PROV Activities, related via \textbf{prov:wasAssociatedWith} (see \figref{prov}).
  \item \textbf{Plan Beliefs} are specializations of \textbf{prov:Entity}.  They support tasks with \textbf{prov:used} and they are realized by tasks with \textbf{prov:wasGeneratedBy}.
  \item \textbf{Information Sources} are specializations of \textbf{prov:Entity}.  They represent sensors and repositories that emit information to derive beliefs and measurements used in the plan, and support beliefs via \textbf{prov:wasDerivedFrom}.
\end{itemize}

As shown in the \figref{prov-to-plan} screenshot, the resulting provenance graph incorporates the information sources (at left) with the goals of the plan (at right), and the dependency network between them.
We use this \shop plan to acquire \textbf{objective1} imagery--with one or more possible planned courses of action--to articulate our approach below.

\begin{figure}
  \begin{center}
  \frame{
  \includegraphics[width=\linewidth]{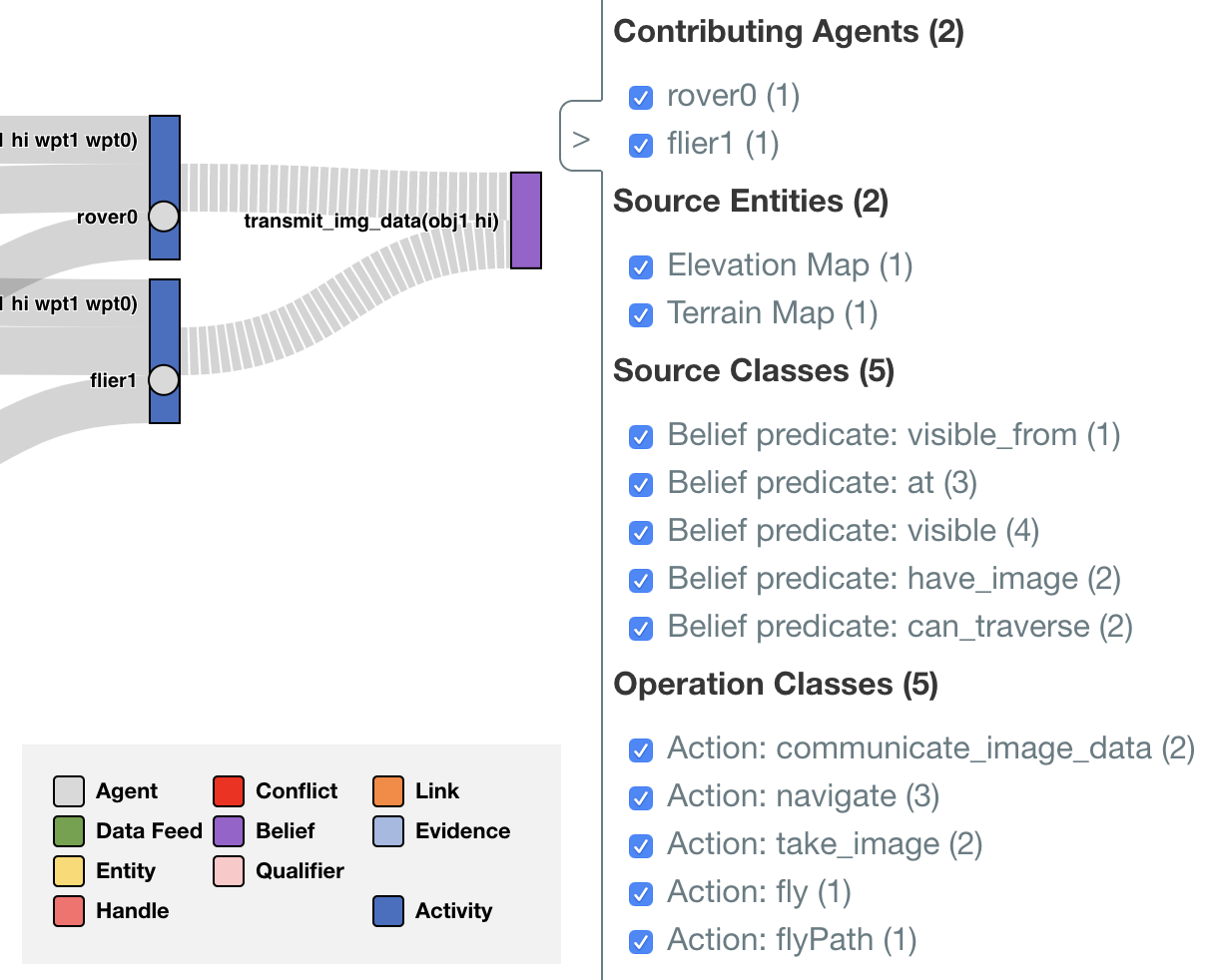}}
  \caption{The index of agents, sources, source classes, and operation classes used in a plan.}
  \label{fig:ss-index}
  \end{center}
\end{figure}

\subsection{Indexing the Dimensions of a Plan}

Given a plan to assess, our provenance system automatically identifies and catalogs the following dimensions of the plan.  These are displayed for user assessment and dynamic interaction, as shown in \figref{ss-index}.
\begin{enumerate}
  \item \textbf{Contributing Agents}: Actors in the plan.
  \item \textbf{Source Entities}: Individual devices or informational resources from which plan-relevant beliefs are derived, such as geolocation, visibility, inventory, and more.
  \item \textbf{Source Classes}: General categories of information across beliefs and information sources.
  These may include information sources 
  or belief predicates, as shown in \figref{ss-index}.
  \item \textbf{Operation Classes}: General categories of activities, spanning potentially many planned activities.
  In \figref{ss-index}, we catalog classes of actions.
\end{enumerate}
\noindent
Cataloging plan nodes along these dimensions allows our approach to instantaneously identify, emphasize, or refute nodes along these dimensions to support explanation.
These elements are identified by mining the predicates and sources of the plan, but could also be informed by the planner's model, in future work.

We use an algorithm similar to assumption-based truth-maintenance and explanation-maintenance systems \cite{forbus1993building,friedman2018csj} to compute the \emph{environment} of all nodes (\ie{}, planned action or belief) in the provenance graph.
The algorithm traverses backward exactly once from all sink nodes, so it reaches each node $m$ in the provenance graph and computes its environment $E(m) = \{S_1, ..., S_n\}$, a disjunction of sets ($S_i$) of \emph{assumptions}, where any $S_i \in E(m)$ is sufficient to derive (\ie{}, believe, achieve, or enact) $m$, and where the assumptions correspond to root nodes in the provenance graph.
The algorithm attends to the AND- and OR-like links listed in \figref{prov-to-plan} to properly encode disjunctive derivation trees.
This compact index answers questions of necessity and sufficiency in constant time.

The joint indexing of plan nodes by the four above dimensions and by their environments allows the provenance analysis system to identify abstract classes of sources and operations that contribute to it, and that it contributes to.
We leverage these indices to help explain the plan in context, as we describe below.

\begin{figure*}[t]
  \begin{center}
  \frame{
  \includegraphics[width=\textwidth]{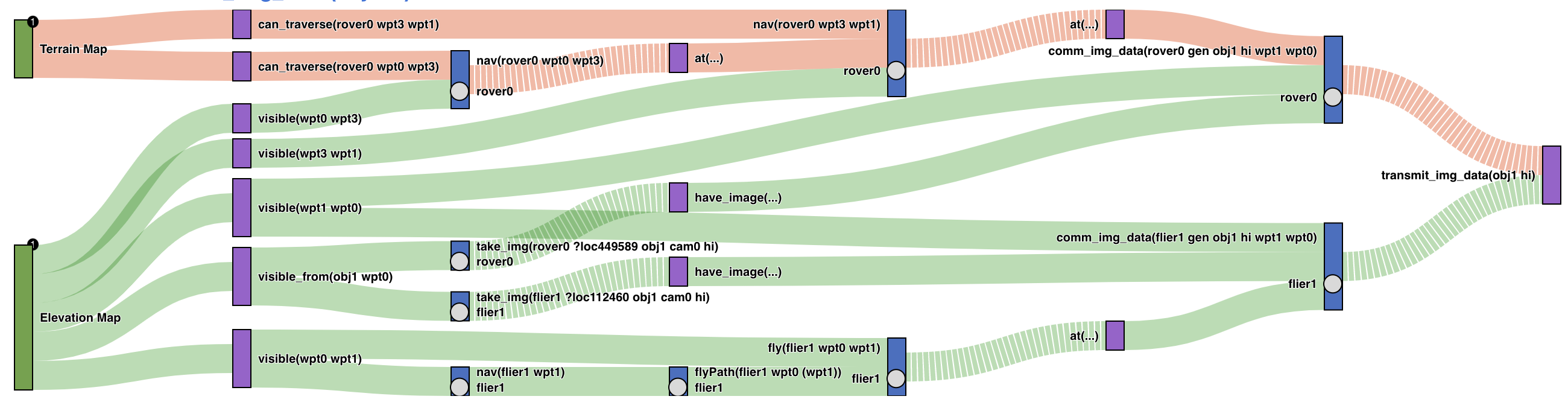}}
  \caption{The \figref{prov-to-plan} plan to acquire imagery of \textbf{objective1}, with moderately low (0.20) confidence ascribed to the \textbf{Terrain Map} and moderately high (0.80) confidence ascribed to the \textbf{Elevation Map}.}
  \label{fig:ss-compare}
  \frame{
  \includegraphics[width=\textwidth]{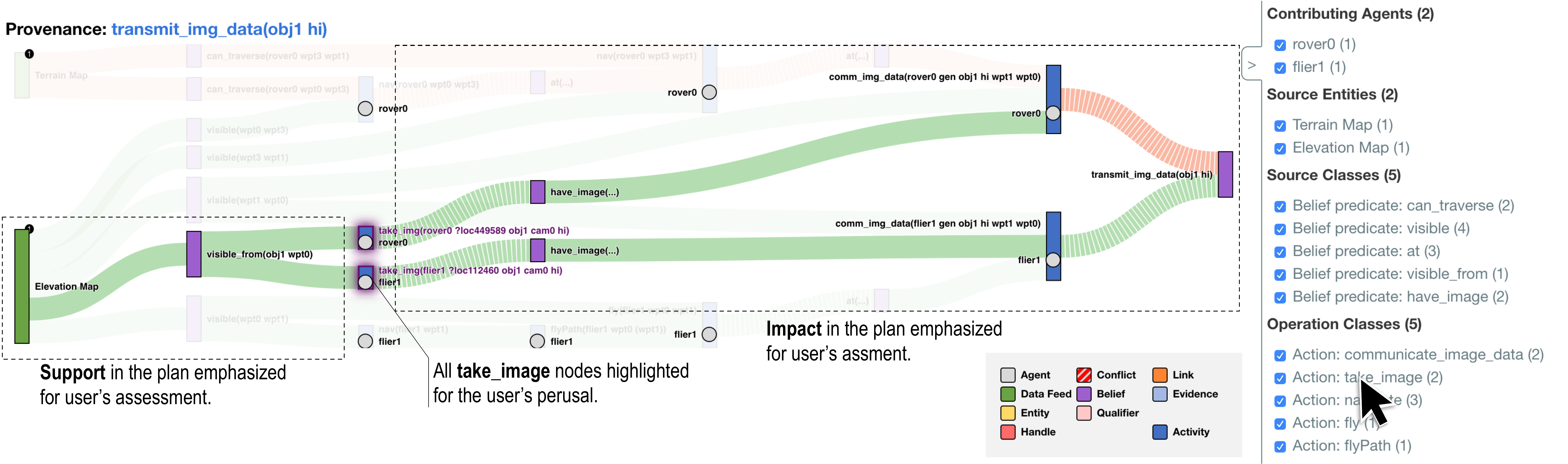}}
  \caption{Viewing the relevance and reachability of the \textbf{take-image} action, including the sources it relies on, and the impact on downstream actions.}
  \label{fig:ss-impact-take-image}
  \end{center}
\end{figure*}

\subsection{Visualization Environment}
\label{sec:viz}

Our visualization environment is a graphical display within a larger web-based platform for human-machine collaborative intelligence analysis.
At any time, the user may select one or more elements from diagrams or listings and peruse its full provenance.

A web service traverses the knowledge graph to retrieve the full provenance for the desired belief(s) and all relevant appraisals, and then sends it to the client.
The client's provenance visualizer uses D3.js, as shown in the \figref{prov-to-plan} and \figref{ss-index} screenshots, to implement the rendering, refutation, emphasis, and propagation effects described below, operating over the PROV and DIVE representations.


\section{Assessing Explainability of our Approach}

The majority of prior work on explainable planning focuses on inspecting algorithms (\ie{}, explicating the decision-making process), synchronizing mental models (\eg{}, because the user views the problem differently than the planner), and improving usability (\eg{}, making complex plans more interpretable) \cite{ChakrabortiEtAlIJCAI2020} and assumed fixed background domain knowledge.  In contrast, our provenance-based approach treats the plan as a tripartite (\textbf{Agents}, \textbf{Entities}, and \textbf{Activities}) dependency graph.
This adds connections among the plan's beliefs and goals (PROV entities), actions (PROV activities), and actors (PROV agents) via type-specific dependency relations.
The plan's provenance graph connects to other provenance information (if available), including belief derivations (\eg{}, describing how initial state beliefs were inferred, as in \figref{prov-to-plan}), agent descriptions, and sensor descriptions (\eg{}, including reliability information), which comprise a larger global provenance graph.
This complements previous explainable planning work with additional decision-relevant information and thereby new explanation capabilities.

\subsection{Explanation in Information Analysis}

We review questions from information analysis that are relevant but under-explored for automated planning, especially when a plan's world state is derived and supported by diverse information.
These questions stem primarily from directives for integrity in intelligence analysis \cite{icd_203,icd_206}, and measurements of rigor in analytic workflows \cite{zelik2010measuring}.
For each question, we briefly note whether our approach addresses it adequately (\textcolor{ForestGreen}{\cmark}) or partially ($\sim$) or whether it is out of scope (\textcolor{BrickRed}{\xmark}).

\vspace{-0.15in}
\paragraph{(\textcolor{ForestGreen}{\cmark}) How reliable is the information supporting this course of action?}
We answer this question of information reliability with graph propagation, using all DIVE \textbf{Appraisal} instances with numerical confidence ratings and propagating them forward to estimate downstream nodes' confidence.
\figref{ss-compare} illustrates the \textbf{Elevation Map} appraised with moderately high (0.80) confidence and the \textbf{Terrain Map} appraised with moderately low (0.20) confidence.
We see that the downstream goal (rightmost node) is supported by two paths of varying estimated confidence, where the low confidence begins at the \textbf{Terrain Map} and flows through the \textbf{rover0} sub-plan.
In the present propagation policy, a conjunction is as reliable as the lowest-confidence upstream input and a disjunction is as reliable as the greatest-confidence source upstream, but Bayesian approaches may also apply here \cite{kuter10using}.

\begin{figure*}[ht!]
  \begin{center}
  \frame{
  \includegraphics[width=\textwidth]{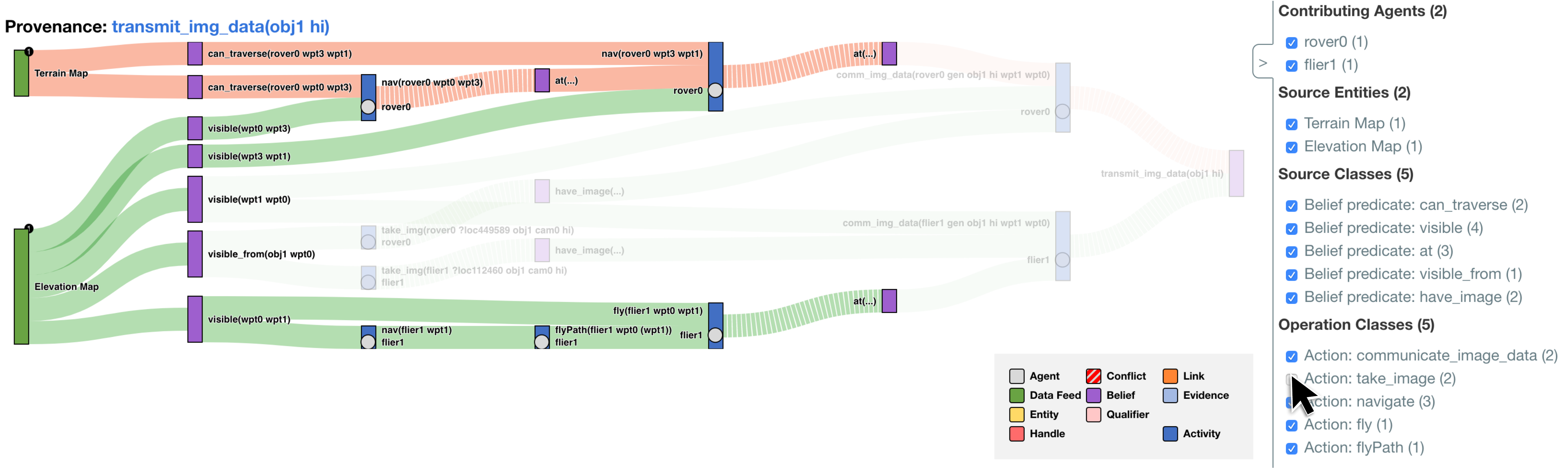}}
  \frame{
  \includegraphics[width=\textwidth]{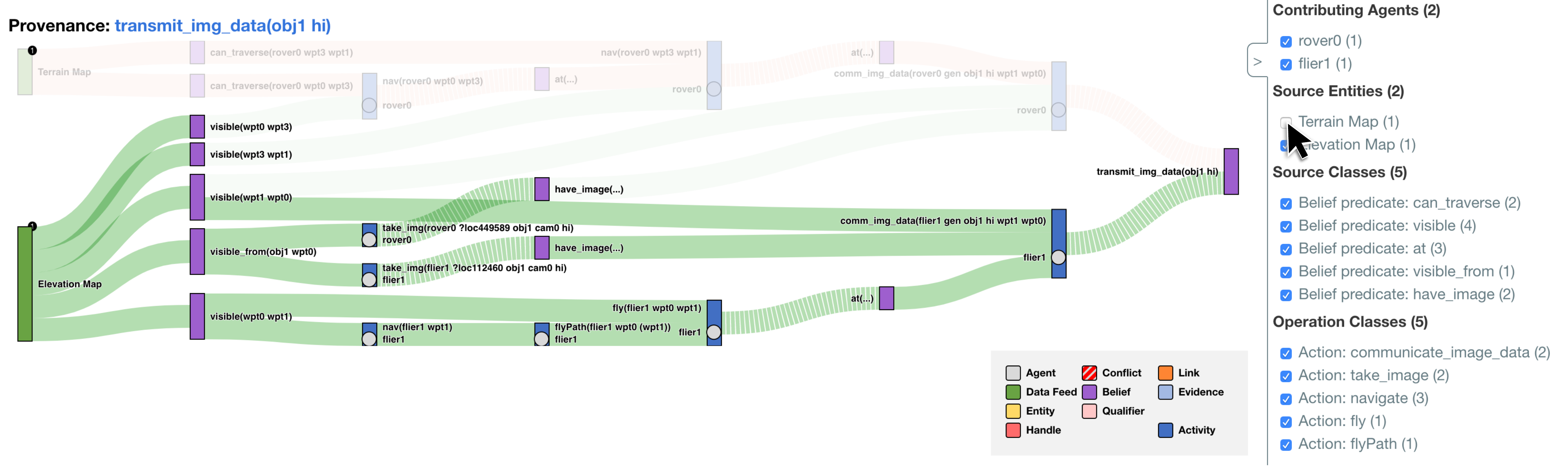}}
  \frame{
  \includegraphics[width=\textwidth]{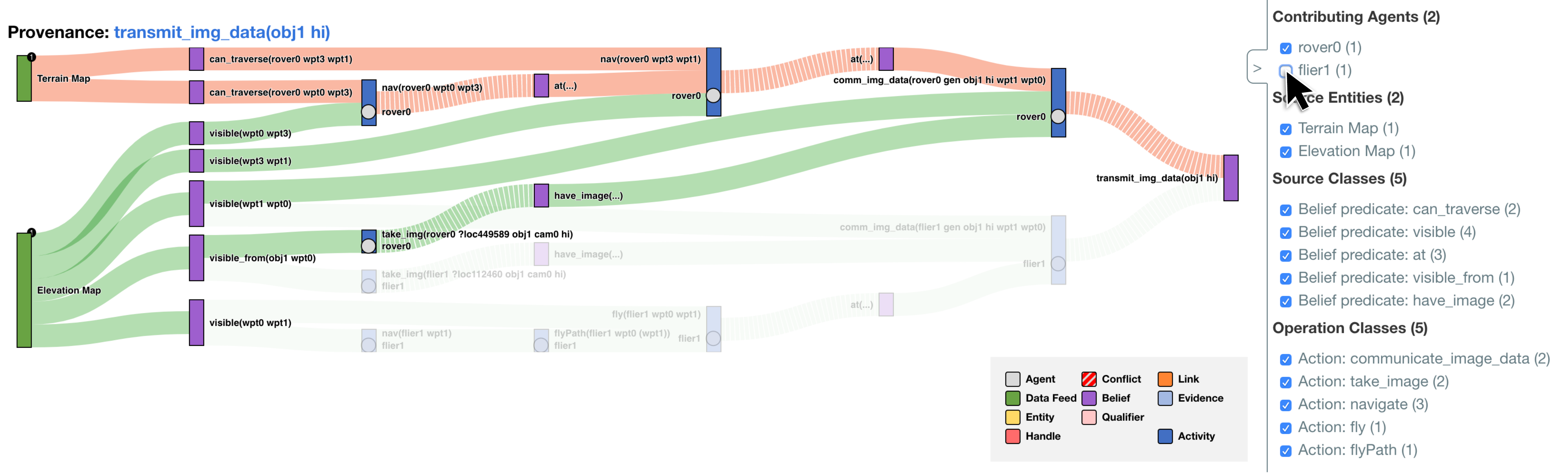}}
  \caption{Counter-factually refutation of: a class of operations \textbf{take\_image} (top); an information source \textbf{Terrain Map} (middle); and an entire agent \textbf{flier1} (bottom) from the plan.
  Refuting elements allows us to see the impact on downstream goals and actions, and the confidence of the information supporting them.}
  \label{fig:ss-refutation}
  \end{center}
\end{figure*}

\vspace{-0.15in}
\paragraph{(\textcolor{ForestGreen}{\cmark}) What information sources, sensors, or actors are pertinent to this [class of] belief or action?}
Our system answers this question of \textit{information support} using the pre-computed environment (described above) to identify all upstream necessary and sufficient nodes in constant time.
The \figref{ss-impact-take-image} screenshot shows the effect of hovering over the \textbf{take\_image} action in the right-hand panel: the system (1) identifies all nodes catalogued with that action (highlighted with purple glow in \figref{ss-impact-take-image}), and then (2) de-emphasizes all nodes and paths that are not pertinent, so all relevant supporting nodes (upstream of the \textbf{take\_image} nodes) are available for assessment.
We see that all \textbf{take\_image} actions rely solely on (1) a belief about \textbf{objective1} visibility from \textbf{waypoint0} and (2) a high-confidence information source.

\vspace{-0.15in}
\paragraph{(\textcolor{ForestGreen}{\cmark}) How far has this belief/agent/information source influenced my plan?}
Our approach answers this \textit{impact assessment} question using belief environments: the impact of a belief, agent, or information source $m$ in the provenance graph is the set of elements with $m$ in any subset of their environments.
The impact of the \textbf{take\_image} nodes is shown downstream of those nodes in the \figref{ss-impact-take-image} screenshot: the \textbf{take\_image} actions directly impact the communication of image data, in both sub-plans, thereby indirectly impacting the rightmost goal along both avenues.

\vspace{-0.15in}
\paragraph{(\textcolor{ForestGreen}{\cmark}) How necessary are these sources, beliefs, actions, or actors for an action or goal?}
This is known as \textit{sensitivity analysis}, and is answerable using environments, as defined above.
Given an element $m$, we can answer whether one or more other elements $N$ are necessary by computing $m$'s environment contracted by $N$:

\newcommand{\ssep}{:}

\[
E(m)/N = \{\, S \in E(m) \ssep N \cap S = \emptyset \,\}
\]

\noindent
If $E(m)/N = \emptyset$, at least one element in $N$ is necessary for $m$.
This allows us to interactively \textit{refute} elements in the provenance graph and observe the downstream effects, answering counter-factual ``\textit{what-if}'' questions about the necessity of information and actors in the plan.

Our system supports sensitivity analyses via dynamic \emph{refutation} as shown in \figref{ss-refutation}: the user may refute a class of elements (\figref{ss-refutation}, top); information sources (\figref{ss-refutation}, middle); agents (\figref{ss-refutation}, bottom); or any individual node.
The system contracts nodes' belief environments, as described above, to identify downstream that have lost all support.
Note that the downstream goal is still reachable in two of these \figref{ss-refutation} refutations; however, the confidence of the goal varies depending on which element we refute.

\vspace{-0.15in}
\paragraph{(\textcolor{ForestGreen}{\cmark}) What assumptions are necessary or sufficient to hold this belief or apply this planned action?}
Deriving beliefs from information sources often requires making some assumptions.
For instance, using a rover's GPS sensor to measure its position assumes that \textit{the GPS sensor is on the rover}.
This assumption affects the integrity of all downstream beliefs and planned actions that rely directly or indirectly on positional data.

As with numerical confidence, we express assumptions using DIVE \textbf{Appraisal} instances related to the relevant elements (\eg{}, a GPS sensor).
For any node $m$, we compute the set of necessary and sufficient upstream assumptions as the set of explicit assumptions on the necessary and sufficient nodes in $E(m)$.

\subsection{Explanation in Automated Planning}

We consider explainable planning questions from
\citeauthor{MariaFoxExplainablePlanning2017} \shortcite{MariaFoxExplainablePlanning2017} in relation to our approach:
\hide{
\begin{enumerate*}[label=(\arabic*)]
\item \emph{Why did you do that?}
\item Why didn't you do something else?
\item \emph{Why is what you propose to do more efficient/safe/cheap than something else?}
\item Why can't you do that?
\item \emph{Why do I need to replan or repair the plan at this point?}
\item \emph{Why do I not need to replan or repair the plan at this point?}
\end{enumerate*}
}

\vspace{-0.15in}
\paragraph{(\textcolor{ForestGreen}{\cmark}) Why did you do that?}
Given any action, our approach uses the provenance structure to identify source nodes (\ie{}, information sources), sink nodes (\ie{}, goals), and intermediate nodes (\ie{}, beliefs and other actions) that explain upstream information support and downstream depenedencies.  Through the interface, one can simply hover
their mouse over the inquired action and view these relationships from the provenance structure (see \figref{ss-impact-take-image}).  The
upstream information support indicates a justification for how the decision contributes to the goals, and
the downstream dependencies provide reasons for why the decision was possible to make (compared to other
potential decisions that could also achieve the goal).

While this can be as simple as visualizing causal
links of preconditions and effects, the reliability of information can also play a role.  The inquired action
may involve entities with greater DIVE \textbf{Appraisal}, for example.

\vspace{-0.15in}
\paragraph{(\textcolor{BrickRed}{\xmark}) Why can't you do that?}

This question concerns an action that was \emph{not} included in the plan, and the analysis in this work is limited to analyzing components within the plan (\ie{}, only actions emitted by the planner).
The provenance
structures are constructed based on the \emph{outcome of the planning process}, which is an annotated HTN
without vestigial structures from actions that were not selected for the returned plan.
Therefore, this question is out of scope for our provenance-based analysis.

\vspace{-0.15in}
\paragraph{(\textcolor{BrickRed}{\xmark}) Why didn't you do something else?}

This question frames a planned action against the space of other, unplanned actions.  Thus it is also out
of scope for similar reasons to the above explainability question.  Without the context of an unplanned (novel) action, such comparisons between actions cannot be made.

\vspace{-0.15in}
\paragraph{($\sim$) Why is that more efficient/safe/cheap than something else?}
Our provenance-based approach propagates confidence---or alternatively, source reliability or operational risk---downstream through the provenance graph, allowing upstream agents, beliefs, and information sources to color downstream actions and beliefs in the plan.
This estimation of downstream confidence and risk (as an inverse of ``safe,'' per the question) allows us to compare alternatives across numerical measures.
This does not fully address the question, since propagating confidence does not explain resource costs and efficiency.

\vspace{-0.15in}
\paragraph{($\sim$) Why do I [not] need to replan or repair the plan at this point?}

This extends to specific questions about plan robustness such as, ``What can go wrong with this plan, and why?'' \eg, ``what will happen if this rover breaks down?''
Connecting
the rover to actions and goals that involve it enable the planning system to explain the overall impacts
of such a query, rather than simply identify the chain of broken causal links in a single plan instance
\cite{Bercher14HybridPlanningApplication}.

It is trivial to reassign a DIVE \textbf{Appraisal} of an entity,
since the provenance structures do not change: the new values propagate after updating the confidence and reliability of the remaining plan components.  Hence reducing the reliability of a
rover that seems likely to break down will downgrade the estimated confidence in the portion of the plan that the rover supports.
Similarly, dynamically refuting the unreliable rover, as illustrated in \figref{ss-refutation}, will instantly remove elements of the plan that rely on it.

If there are still sufficient paths to the goal condition---or paths that are of the desired confidence---then the plan is robust enough to
address the inquired failure points, and it does not require revision.  Alternatively, if the remaining paths to the goal are not of a desired confidence, then these refuted elements (and the degraded paths) explain why revising the plan is necessary.

\vspace{-0.15in}
\paragraph{}
\noindent{}


\section{Related Work}
\label{sec:related}

In the AI planning community, it has shown that it is possible to formalize ``model synchronization'' as a meta-search problem, where traditional search and classical planning algorithms explore explanations with differing properties \cite{chakraborti1802plan,chakraborti2018explicability}. One key insight of model synchronization is that explanations are generally needed to identify mismatches in planning models produced by different information sources. This is critically important when the distance between different descriptions of a planning domain   cannot capture a cohesive model sufficiently. While explanations can certainly deviate from our actual methods of decision making \cite{klein08} they nevertheless represent how humans are trained and acculturated to providing rationalizations for our decision making. In that sense, we believe this line of work is complementary to our approach in this paper: two approaches can be combined in order to formalize and reason about properties such as social trust, analytic trust, communication frequencies, and others. This approach can balance the trade-offs between explicability and explanations for social interactions. In particular, an “optimal” AI agent might generate an estimate of the state of the world that is inexplicable to humans and model synchronization and provenance tracing will enable an AI agent to choose a less optimal model of the state that would enable an easier explanation to the human users and is close to (but not the same as) the AI agent’s actual domain models \cite{chakraborti2018explicability}.

Provenance-tracking is well-established as a practical tool across source domains \cite{gehani2010mendel,pasquier2016information} for decision support \cite{singh2018decision}, complex multi-agent workflows \cite{toniolo2015supporting,friedman_tapp_2020}, and lineage-tracking for databases \cite{benjelloun2008databases}.
These previous works have not been applied to the domain of planning, so we believe this work is the first to investigate the explainability of automated planning using provenance.

Label propagation has been used to detect persistent security threats in real time \cite{han2020unicorn} by propagating information through network flows.
We have previously used label propagation for
plan recognition with support for refutation, similar to what we demonstrate here \cite{primrose2018}, but this previous approach did not use formal provenance notation or operate on forward-generated plans.

\section{Conclusions}
\label{sec:conclusions}

This paper presented a provenance-based approach for improving the explainability of plans.
Our approach (1) extends the \shop HTN planner to generate dependency information, (2) transforms the dependency information into an established PROV-O representation, and (3) uses graph propagation and TMS-inspired algorithms support dynamic and counter-factual assessment of information flow, confidence, and support.

We qualified our approach's explanatory scope with respect to explanation targets from the automated planning literature \cite{MariaFoxExplainablePlanning2017} and the information analysis literature \cite{icd_203,icd_206,zelik2010measuring}.
We demonstrated that our approach answers questions of pertinence, sensitivity, risk, assumption support, diversity of evidence, and relative confidence.
Our approach is limited to explaining elements of the plan itself: it does \emph{not} explain why a given action was not planned or whether an action is plannable via the planner's internal model.

Our provenance approach might be able to help explain ``the road not taken'' if the planner represents decision points and constraints in the dependency-based plan.
This would not enable complete \emph{``what-if''} hypothetical explanations, but it would explain local rationale for planning decisions within context.

This paper used simple example plans.
Our underlying graph propagation and TMS algorithms easily scale to larger datasets, and they can execute incrementally when graphs (\ie{}, automated plans) are revised online \cite{forbus1993building}.
However, we face a non-computational scalability problem of user experience: the UI to display the associated provenance cannot intuitively display full plans with hundreds of nodes without graph summarization and graph filtering, which is one avenue of future work.

\subsection{Future Work}
This work demonstrates the explanatory value of provenance for analysis \emph{after} the planning process.
We see value in integrating these provenance-based analyses \emph{online} into a continuous and dynamic planning environment, interleaving  provenance analysis and planning.
Some possibilities for using provenance while planning include: heuristic guidance (\eg, preferring choices based on higher-confidence information); 
guiding contingency planning (\eg{}, prepare for more likely nondeterministic outcomes based on reliability of sensors, information sources, \etc);
or to 
plan repair (\eg{} triggering the planner to make revisions when provenance changes for the worse).

Furthermore, as a source of explainability to other agents, there is a potential for novel uses in multi-agent planning scenarios such as decentralized planning (\eg{}, evaluating other agents' performance to assess reliability of their action outcomes and relayed information)
and mixed-initiative planning (\eg{}, using the interface to detail the current plan's provenance and receive iterative changes from the user to parameters such as DIVE \textbf{Appraisals}).



\section*{Acknowledgments}

This work was funded primarily by a SIFT Internal R\&D project.
We thank Kanna Rajan for his suggestions.

\bibliographystyle{aaai}
{\small
\bibliography{proj7,sensemaking,explainable-planning,main}
}
\end{document}